\documentclass[10pt,journal,compsoc]{IEEEtran}

\ifCLASSOPTIONcompsoc
  \usepackage[nocompress]{cite}
\else
  \usepackage{cite}
\fi

\ifCLASSINFOpdf
  \usepackage[pdftex]{graphicx}
\else
  \usepackage[dvips]{graphicx}
\fi

\usepackage{booktabs} 
\usepackage{float}
\usepackage{pifont}
\usepackage{color}
\usepackage[ruled,vlined]{algorithm2e}
\usepackage{algorithmic}
\usepackage{multirow}
\usepackage{amsmath,amssymb}
\usepackage{bbm}
\usepackage{mathrsfs}
\usepackage{array,multirow}
\DeclareMathOperator{\E}{\mathbb{E}}
\newcolumntype{C}[1]{>{\centering\arraybackslash}m{#1}}
\newcommand{\xmark}{\ding{55}}%

\DeclareMathOperator*{\argmin}{argmin}
\begin{document}

\title{Defending against Adversarial Attack towards Deep Neural Networks via Collaborative Multi-Task Training}

\author{Derui~(Derek)~Wang,
        Chaoran~Li,
        Sheng~Wen,
        Surya~Nepal,
        and~Yang~Xiang
\IEEEcompsocitemizethanks{\IEEEcompsocthanksitem {D. Wang, C. Li, S. Wen, and Y. Xiang} are with School of Software and Electrical Engineering, Swinburne University of Technology, Hawthorn, VIC 3122, Australia;\protect\\
E-mail: \{deruiwang, chaoranli, swen, yxiang\}@swin.edu.au;
\IEEEcompsocthanksitem D. Wang, C. Li, and S. Nepal are with Data 61, CSRIO, Australia;}

\thanks{The manuscript was submitted on October 15, 2018, and was accepted on July 15, 2020.}}

\markboth{IEEE Transactions on Dependable and Secure Computing}%
{Shell \MakeLowercase{\textit{et al.}}: Bare Demo of IEEEtran.cls for Computer Society Journals}

\IEEEtitleabstractindextext{%
\begin{abstract}
Deep neural networks (DNNs) are known to be vulnerable to adversarial examples which contain human-imperceptible perturbations. A series of defending methods, either proactive defence or reactive defence, have been proposed in the recent years. However, most of the methods can only handle specific attacks. For example, proactive defending methods are invalid against grey-box or white-box attacks, while reactive defending methods are challenged by low-distortion adversarial examples or transferring adversarial examples. This becomes a critical problem since a defender usually does not have the type of the attack as a priori knowledge. Moreover, existing two-pronged defences (\textit{e.g.}, MagNet), which take advantages of both proactive and reactive methods, have been reported as broken under transferring attacks. To address this problem, this paper proposed a novel defensive framework based on collaborative multi-task training, aiming at providing defence for different types of attacks. The proposed defence first encodes training labels into label pairs and counters black-box attacks leveraging adversarial training supervised by the encoded label pairs. The defence further constructs a detector to identify and reject high-confidence adversarial examples that bypass the black-box defence. In addition, the proposed collaborative architecture can prevent adversaries from finding valid adversarial examples when the defence strategy is exposed. In the experiments, we evaluated our defence against four state-of-the-art attacks on $MNIST$ and $CIFAR10$ datasets. The results showed that our defending method achieved up to $96.3\%$ classification accuracy on black-box adversarial examples, and detected up to $98.7\%$ of the high confidence adversarial examples. It only decreased the model accuracy on benign example classification by $2.1\%$ for the $CIFAR10$ dataset.
\end{abstract}

}

\maketitle

\IEEEdisplaynontitleabstractindextext

%
\IEEEpeerreviewmaketitle

\IEEEraisesectionheading{\section{Introduction}\label{sec:introduction}}

%
%
%
%
\IEEEPARstart{D}{eep} neural networks (DNNs) have achieved remarkable performance on various tasks, such as computer vision, natural language processing and data generation. However, DNNs are vulnerable towards adversarial attacks which exploit imperceptibly perturbed examples to fool the neural networks \cite{Szegedy2013Intriguing}. For instance, deliberately crafted adversarial examples can easily deceive DNN-based systems such as hand-writing recognition \cite{papernot2016limitations}, face detection \cite{li2015convolutional}, and autonomous vehicle \cite{girshick2016region}, making the models generate wrong outputs. The adversarial examples are even possible to trigger catastrophic consequences, such as causing accident originated by faulty object detection in autonomous vehicles \cite{lu2017adversarial, wang2019daedalus}. Considering the blooming DNN-based applications in modern electronic systems, as well as it will likely be in the future applications, proposing effective defensive methods to defend DNNs against adversarial examples has never been this urgent and critical.

Adversarial attacks against DNN-based systems can be categorised into three types based on the a priori knowledge that attackers have: 1) black-box attacks, 2) grey-box attacks, and 3) white-box attacks. Currently, there are also a series of defending methods proposed, which can be broadly divided into proactive defence (\textit{e.g.}, \cite{Gu2015Towards,papernot2016distillation,cisse2017parseval,shaham2018understanding}) and reactive defence (\textit{e.g.}, \cite{xu2017feature,Feinman2017,grosse2017statistical,lu2017safetynet,metzen2017detecting}). Proactive defence increases the robustness of victim models against adversarial examples, while reactive defence detects the adversarial examples from the model inputs. However, they all have limitations. For example, current countermeasures can only counter attackers in specific scenarios.

For proactive defending methods, they focus on transferring attacks launched in black-box settings. However, they do not work in grey-box and white-box scenarios since this type of defence relies on model parameter regularisation and robust optimisation to mitigate the effects of adversarial examples. They become invalid once the parameters or the defending strategies are known to attackers. Moreover, this type of defence can be bypassed by using high-confidence adversarial examples (\textit{e.g.}, Carlini$\&$Wagner attack \cite{carlini2017}), even in black-box settings. This type of defence is also not resilient to the attacks that exploit input feature sensitivity (\textit{e.g.}, Jacobian matrix based attacks \cite{papernot2016limitations}). Instead of passively strengthening models' robustness, reactive defending methods can capture adversarial examples that have higher attacking confidence and distortion \cite{xu2017feature,Feinman2017, grosse2017statistical, lu2017safetynet, metzen2017detecting}. However, these methods have been demonstrated to be vulnerable to specific transferring attacks \cite{carlini2017adversarial}.

Methods employed by an attacker are usually hidden to a defender in practice, which makes the selection of a proper defence be very challenging. We also cannot simply ensemble the aforementioned defending methods together to form an intact defence, as attackers can compromise each defending method one by one \cite{he2017adversarial}. Magnet \cite{meng2017magnet} provides a two-pronged defence that does not require too much a priori knowledge on the type of the coming attack. However, this method has been reported to be broken by transferring attack directed from substitute autoencoders \cite{carlini2017magnet}. A recent work suggests using Local Intrinsic Dimensionality (LID) of an example to investigate whether the example is adversarial or not\cite{ma2018characterizing}. However, LID has been proved as vulnerable to high confidence adversarial examples \cite{obfuscated-gradients}. DeepFense is proposed as another defence that can be generalised to unknown attacks\cite{rouhani2018deepfense}. However, it requires extensively hardware acceleration, which makes it difficult to deploy to low-resource devices. So far, there is no easy-to-deploy and intact defence that can be generalised to different attacks.

In this paper, we propose a well-rounded defence that increases the robustness of neural networks to low-confidence black-box attacks and detects high-confidence black-box adversarial examples at a high accuracy. Moreover, our proposed defence can prevent an adversary from finding adversarial examples in grey-box scenario. We provide the detailed definition of black-box and grey-box scenarios in Section \ref{S::Threat_Model}. Our method first introduces adversarial training with robust label pairs to tackle black-box attack. Then it employs a multi-task training technique to construct an adversarial example detector. The proposed method is able to tackle both the transferring attack launched in the black-box setting and the generation of adversarial examples based on the targeted model in the grey-box setting. The main contributions of the paper are summarised as follows:

\begin{itemize}
\item\textit{We introduced a novel collaborative multi-task training framework as a defence to invalidate transferring adversarial examples;}
\item\textit{This defence uses data manifold information to detect high-confidence adversarial examples crafted in grey-box/black-box settings;}
\item\textit{The proposed defence can prevent an adversary from searching valid adversarial examples using the targeted model in grey-box settings;}
\item\textit{The proposed defence is resilient to the transferring attack which breaks the previous two-pronged defence (\textit{e.g}. MagNet);}
\item\textit{We carry out both empirical and theoretical studies to evaluate the proposed defence. The experimental results demonstrate that our defence is effective against adversaries with different prior knowledge.}
\end{itemize}

The paper is organised as follows: Section \ref{problem_statement} describes the state-of-the-art attacks and clarifies the problem statement and our contributions. Section \ref{method} presents our detailed approach. Section \ref{results} presents the evaluation of our approach. Section \ref{justification} provides an analysis on the mechanism of the defence. Section \ref{related_work} presents a conclusion on the existing attacks and the defensive methods. Section \ref{discussion} discusses the remaining unsolved problems of the existing attacks and defences, as well as the possible further improvements of the defence. Section \ref{conclusion} summaries the paper, and proposes the future works.

\section{Primer}\label{problem_statement}
\subsection{Adversarial attacks}
We first introduce five representative attacks as background knowledge. Supposing the DNN model is equal to a non-convex function $F$. In general, given an image $x$ along with the rightful one-hot encoded label $y_{true}$, an attacker searches for the adversarial example $x_{adv}$. 

\subsubsection{FGSM}
Fast gradient sign method (FGSM) is able to generate adversarial examples rapidly \cite{Goodfellow2014Explaining}. FGSM perturbs an image in the image space towards gradient sign directions. FGSM can be described using the following formula:

\begin{equation}
x_{adv} \leftarrow x + \epsilon sgn(\bigtriangledown_{x}{L(F(x),y_{true})})
\end{equation}

Herein $L$ is the loss function (a cross-entropy function is typically used to compute the loss). $F(x)$ is the softmax layer output from the model $F$. $\epsilon$ is a hyper-parameter which controls the distortion level on the crafted image. $sgn$ is the sign function. FGSM only requires gradients to be computed once. Thus, FGSM can craft large batches of adversarial examples in a very short time.

\subsubsection{IGS}
Iterative gradient sign (IGS) attack perturbs pixels in each iteration instead of a one-off perturbation \cite{kurakin2016adversarial}. In each round, IGS perturbs the pixels towards the gradient sign direction and clip the perturbation using a small value $\epsilon$. The adversarial example in the $i$-th iteration is stated as follows:

\begin{equation}
x^i_{adv} = x^{i-1}_{adv}-clip_{\epsilon}(\alpha \cdot sgn(\bigtriangledown_{x}{L(F(x^{i-1}_{adv}),y_{true})}))
\end{equation}

Compared to FGSM, IGS can produce an adversarial example with a higher mis-classification confidence.

\subsubsection{Deepfool}
Deepfool is able to generate adversarial examples with minimum distortion on original images \cite{Moosavidezfooli2016DeepFool}. The basic idea is to search for the closest decision boundary and then perturb $x$ towards the decision boundary. Deepfool iteratively perturbs $x$ until $x$ is misclassified. The modification on the image in each iteration for binary classifier is calculated as follows:

\begin{equation}
r_i \leftarrow -\frac{F(x)}{\parallel{\bigtriangledown{F(x)}}\parallel^{2}_{2}} \bigtriangledown{F(x)}
\end{equation}

Deepfool employs the linearity assumption of the neural network to simplify the optimisation process. We use the $L_{\infty}$ version of Deepfool in our evaluation.

\subsubsection{JSMA}
Jacobian-based saliency matrix attack (JSMA) iteratively perturbs important pixels defined by the Jacobian matrix based on the model output and input features \cite{papernot2016limitations}. The method first calculates the forward derivatives of the neural network output with respect to the input example. The adversarial saliency map demonstrates the most influential pixels which should be perturbed. Based on two versions of the saliency map, attacker can increase the value of the influential pixels in each iteration to generate targeted adversarial examples, or decrease pixel values to get non-targeted examples.

\subsubsection{Carlini$\&$Wagner $L_{2}$}
This method has been reported to be able to make defensive distillation invalid \cite{carlini2017}. This study explored crafting adversarial examples under three distance metrics (\textit{i.e.} $L_0$, $L_2$,and $L_{\infty}$) and seven modified objective functions. We use Carlini$\&$Wagner $L_{2}$, which is based on the $L_2$ metric, in our experiment. The method first redesigns the optimisation objective $f(x_{adv})$ as follows:

\begin{equation}
f(x_{adv})=max(max\{Z(x_{adv})_{i}:i\ne l\}-Z(x_{adv})_{l},-\kappa)
\end{equation}

where $Z(x_{adv})$ is the output logits of the neural network, and $\kappa$ is a hyper-parameter for adjusting adversarial example confidence at the cost of enlarging the distortion on the adversarial image. Then, it adapts L-BFGS solver to solve the box-constraint problem:

\begin{equation}
\min_{\delta}{\|\delta|_{2}^{2}+c\cdot f(x+\delta)}\\
s.t.\ \ x+\delta\in [0,1]^{n}
\end{equation}

Herein $x+\delta=x_{adv}$. The optimisation variable is changed to $\omega: \delta=\frac{1}{2}{tanh(\omega)+1}-x$. According to the results, this method has achieved $100\%$ attacking success rate on the distilled networks in a white-box setting. By changing the confidence, this method can also have targeted transferable examples to perform a black-box attack.

\subsection{Threat model} \label{S::Threat_Model} 
In the real-world cases, an adversary normally do not have the parameters or the architecture of a deep learning model, since the model is well-protected by a service provider. Moreover, in prior works \cite{carlini2017,carlini2017adversarial}, it is recommended that the robustness of a model should be evaluated by transferring adversarial examples. Otherwise, attackers can use an easy-to-attack model as a substitute to break the defence on the oracle model. Therefore, as our first threat model, we assume that the adversary is in a black-box setting. Second, in some cases, the architecture and parameters of the model, as well as the defending mechanism, may be leaked to the attacker. This leads to a grey-box scenario. However, the adversary does not know the parameters of the defence. In an extreme case of white-box scenario, the adversary knows everything about the oracle model and the defence. This is a very strong assumption. Attacks launched in this way are nearly impossible to defend since the attacker can take countermeasure for defence. Therefore, we mainly consider black-box and grey-box threats in our work. We summarise the threat models used in this paper as follows:

\begin{itemize}
\item\textbf{Black-box threats}: an attacker does not know the parameters and architecture of the target model. However, the attacker can train an easy-to-attack model as an substitute to craft adversarial examples, and transfer the examples onto the target classifier (\textit{i.e.} the oracle). The attacker also has a training dataset which has a similar distribution with that of the dataset used to train the oracle. To simulate the worst yet practical case that could happen in the real world, the substitute and the oracle are trained using the same training dataset. However, the attacker knows neither the defensive mechanism, nor the exact architecture and parameters of the oracle.

\item\textbf{Grey-box threats}: an attacker knows the parameters and the architecture of the oracle, as well as the adopted defending method. In this case, the attacker is able to craft adversarial examples based on the oracle instead of the substitute. However, because the parameters of the defensive mechanism are hidden from the attacker, the defence might still be effective.
\end{itemize}

In our work, we assume that the defender has no a priori knowledge pertaining to any of the following questions: 1) what attacking method will be adopted by the attacker, and 2) what substitute will be used by the attacker.

\begin{figure}[t]
\center
\includegraphics[width=1\linewidth]{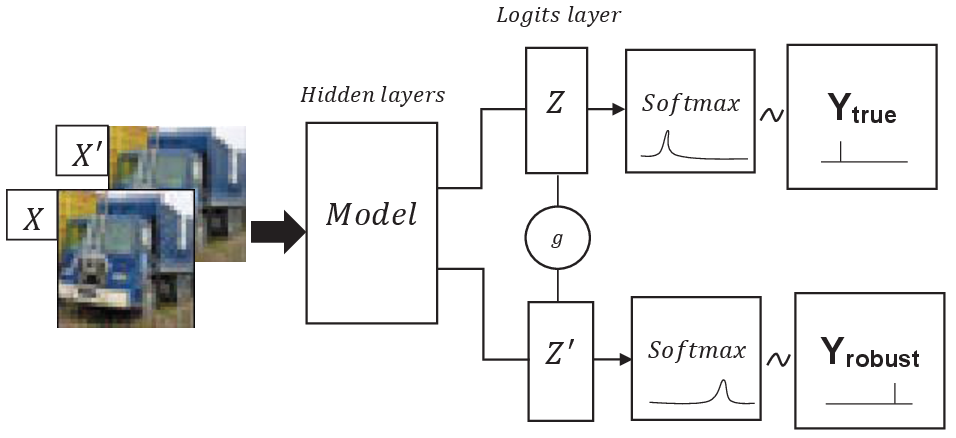}
\caption{The framework of CMT. The training objectives are that, when the incoming input is benign, the outputs from $Y_{true}$ and $Y_{robust}$ will match the pairwise relationship between different classes; When the input is adversarial, the output from $Y_{true}$ stays the same, while the outputs from $Y_{robust}$ will be changed. From another perspective, the output loss with respect to the feature modification is smooth for $Y_{true}$, and is steep for $Y_{robust}$. Moreover, the The collaborative learning by adding a connection between $Z$ and $Z'$ further hardens the model against adversarial example generation in grey-box settings.}
\label{TrainingFramework}
\end{figure}

\section{Design}\label{method} 
We present our defence, namely Collaborative Multi-task Training (CMT), in this section. The intuitions behind our defence are: 1) Our defence framework grows an auxiliary output from the original model. It detects adversarial examples by checking the pairwise relationship of the original output and the auxiliary output against the encoded label pairs. 2) The framework learns smooth decision surfaces for the original output, and steep decision surfaces for the auxiliary output. 3) The smooth manifold learnt for the original output mitigates transferred black-box attacks, while the steep manifold learnt for the auxiliary output ensures that adversarial examples can be detected. 4) The collaborative learning between the two outputs increases the difficulty for generating adversarial examples when an attacker searches adversarial examples given gradients of the victim model. To better encode the label pairs for the original output and the auxiliary output, we first examine the vulnerability of the learnt decision surfaces in a neural network model, such that we can later encode label pairs based on the identified vulnerable decision boundaries. Based on the encoded label pairs, we introduce our collaborative multi-task training framework for defending both black-box attacks and grey-box attacks.

\subsection{Vulnerable decision surface}\label{vds}
In this section, given original data examples of a class, we first identify the corresponding adversarial target class that is usually employed by an adversary. We name the model decision boundary between the original class and the targeted class as a vulnerable decision surface.

A main constraint imposed on adversarial examples is the distortion between a perturbed example and an original example. In the case of non-targeted attack, by using gradient descent to search for adversarial example, the attacker aims to maximise the error of the classification with minimal changes on the example in the feature space. Assuming we have a dataset $D$ and a model $F$ trained on $D$, the decision surfaces of $F$ separate data points belonging to different classes in $D$. According to some previous works, we can find out that, given an example belonging to a certain class, it is easier to target the example to a specific class for the attacker \cite{carlini2017}. This implies that the vulnerable extent of the decision surfaces varies. However, due to the high dimensionality of the features, the vulnerability cannot be measured based on simple metrics (\textit{e.g.}, Euclidean distance). Adversarial examples exist in the local area near the original example. Given a $\delta$-bounded hypersphere centred by the original example, adversarial examples can be found within the hypersphere. Therefore, from another perspective, given a small enough bound $\delta$, adversarial examples that are classified as certain classes may exist outside the hypersphere. It requires more budget on $\delta$ to find adversarial examples in these classes. Moreover, we assume that the distribution of weak adversarial examples is different from that of high-confidence adversarial examples. Henceforth, by statistically learning the difference between the distributions, we may detect high-confidence attacks based on the difference.

In this paper, we conduct a statistical analysis on non-targeted adversarial examples to measure the distribution of example classes in the vicinity of an original example, for each example class. The purpose of our investigation is two-fold: 1) we want to verify whether there is a pattern in the misclassification of a set of adversarial examples; and 2) we also want to verify whether borderline adversarial examples (\textit{i.e.}, low-confidence adversarial examples that reside in the vicinity of benign examples) found by different attacks share similarity in the misclassification pattern. Herein, given original examples in class $i$, we adopt FGSM to generate non-targeted adversarial examples. Subsequently, we record the probability distribution of $i$ being misclassified into other classes $j$. Specifically, we use Eq.\ref{classmap_compute} to estimate a distribution vector $p_i$:

\begin{equation}\label{classmap_compute}
p_i = \frac{1}{N_i}\sum_{n=0}^{N_i}{F(x^i_{adv})}
\end{equation},
wherein, $x^i_{adv}$ is an adversarial example which originally belongs to class $i$. $N_i$ is the number of adversarial examples in the class $i$. $F(x^i_{adv})$ is an output confidence vector from model $F$. $p_i$ yields the expectation of the confidence of an example in $i$ being classified into other classes. Therefore, for each pair of classes, $p_i$ indicates how vulnerable is the learned decision boundary between them. An empirical evaluation is included in Section \ref{cmap}.

\subsection{Encode labels into label pairs}\label{robust_ident}
Following the aforementioned method of vulnerability estimation, we encode labels of the classes in the training dataset to label pairs. To carry out the empirical analysis, we use non-targeted FGSM to generate a set of adversarial examples based on the training dataset. The generated adversarial example set $X_{adv}$ is then fed into the model to get the observation. In the classification results, for a given output label $l^i_{true}$ indicating class $i$, we search the corresponding pairing output label $l^i_{robust}$ by minimising the likelihood of getting the observation of misclassification:

\begin{equation}
l^i_{robust} = \argmin p_i
\end{equation}

Herein, we actually select the least likely class being misclassified into as the robust label paired to $l^i_{true}$. We can then generate the target output distribution $y^i_{robust}$ of the $i$-th example by one-hot encoding $l^i_{robust}$. Following this procedure, we encode the paired-label for each example in the training dataset. The encoding rules between $l^i_{robust}$ and $l^i_{true}$ are saved as a table $Classmap$. In the grey-box setting, this information will be used to access the credibility of an input example.

\subsection{Collaborative multi-task training}
We propose a collaborative multi-task training (CMT) framework in this section. We consider both black-box and grey-box attacks. The proposed general training framework is depicted in Fig.\ref{TrainingFramework}. The framework is trained under a multi-task objective function, which is designed to maximise the divergence between the outputs of adversarial input and benign input. The training process also integrates adversarial gradients in the objective function to regularise the model to defend against Transferring attack.

\subsubsection{Adversarial training for black-box attack}\label{b_version}
According to the label-pair construction method discussed in Section \ref{vds} and Section \ref{robust_ident}, we use the robust label-pairs to conduct the multi-task training \cite{evgeniou2004regularized}. Assuming the original model has the logits layer that has outputs $Z$. Our method grows another logits layer that outputs logits $Z'$ from the last hidden layer. While the softmaxed $Z$ is used to calculate the loss of the model output with the true label $y_{true}$ of the input $x$. The softmax output of $Z'$ is employed to calculate the model output loss with the robust label $y_{robust}$ when $y_{true}$ is given. We also use adversarial examples to regularise the model against adversarial inputs during the training session. The overall objective cost function $J_{obj}$ of training takes the following form:

\begin{align}
J_{obj}=& \alpha J{(x, y_{true})}+\beta J{(x_{adv}, y_{true}))}\\
&+\gamma \mathscr{J'}{(x, x_{adv}, y_{robust})} \nonumber
\end{align}
wherein, $x$ is the benign example. We use adversarial gradients to regularise the model. $x_{adv}$ is the adversarial example produced by the adversarial gradient in the current step. $y_{true}$ is the ground truth label of $x$, and $y_{robust}$ is the most robust label of the current $y_{true}$. $J$ is the cross-entropy cost. $\alpha$, $\beta$, and $\gamma$ are weights adding up to 1.

The first term of the objective function decides the performance of the original model $F$ on benign examples. The second term is an adversarial term taking in the adversarial gradients to regularise the training. The last term moves the decision boundaries towards the most robust class with respect to the current class. As discussed in \cite{Goodfellow2014Explaining}, to effectively use adversarial gradients to regularise the model training, we set $\alpha=\beta=0.4$, and set $\gamma=0.2$. The cost function $\mathscr{J'}$ is the average over the costs on benign examples and adversarial examples:

\begin{equation}
\mathscr{J'}{(x, x_{adv}, y_{robust})} = \frac{1}{2}\{J(x,y_{robust}) + J(x_{adv},y_{robust})\}
\end{equation}
wherein $J(x,y_{robust})$ is the cross-entropy cost, and $J(x_{adv},y_{robust})$ is a negative cross-entropy cost function to maximise the loss of the $y_{robust}$ output, when the input is adversarial.

Once an example is fed into the model, the output through $Z$ and $Z'$ will be checked against the $Classmap$. The example will be recognised as adversarial if the outputs have no match in $Classmap$. Otherwise, it is a benign example, and the output through $Z$ is then accepted as the classification result.

In the black-box setting, the attacker does not know the existence of the defence, the adversarial objective function will only adjust $x$ to produce an example that only changes the output through $Z$ to the adversarial class. However, it cannot guarantee that the output through $Z'$ has the correct encoding rule to the output through $Z$ in the $Classmap$. The grey-box attack is then detected by our architecture.

\subsubsection{An adaptive attack on black-box defence}\label{break_g_defence}
In the grey-box setting, the adversary has access to the model weights and the defensive mechanism. Therefore, adversarial examples can still be crafted against the model under the black-box defence.

In the black-box setting, the attacker does not know the existence of the defence. Henceforth, the adversarial objective function will only adjust $x$ to produce an example that only changes the output through $Z$ to the adversarial class, but cannot guarantee that the output through $Z'$ has the correct mapping relationship to the output through $Z$, according to the $Classmap$. However, in the grey-box setting, when the adversary performs adversarial searching for targeted adversarial examples on the model without the connection $g$, the optimisation solver can find a solution by back-propagating a combined loss function $L(Z(x),Z'(x),t, t')$ as follows:

\begin{equation}
L(Z(x),Z'(x),t, t') = \eta L_{1}(Z(x),t)+ (1-\eta) L_{2}(Z'(x),t')
\end{equation}
wherein $t$ is the targeted output from $Z$, $t'$ is the targeted output from $Z'$. $t$ and $t'$ should be a pair in the $Classmap$. In this paper, we assume that the attacker can feed a large number of adversarial examples through the protected model and find out the paired $t'$ for each $t$. However, this is a very strong assumption, even in the grey-box setting. $\eta$ can be set by the attacker to control the convergence of the solver. The gradients for back-propagating the adversarial loss from logits layer $Z$ and $Z'$ then become:

\begin{equation}
\begin{split}
\frac{\partial L(Z(x),Z'(x),t, t')}{\partial x} &= \eta\cdot\frac{\partial L_{1}(Z(x),t)}{\partial Z(x)}\frac{\partial Z(x)}{\partial x}+\\&(1-\eta)\cdot\frac{\partial L_{2}(Z'(x),t')}{\partial Z'(x)}\frac{\partial Z'(x)}{\partial x}
\end{split}
\end{equation}

Therefore, it can be observed that the solver can still find an adversarial example by using a simple linear combination of the adversarial losses in the objective functions, in the grey-box setting. The detection method used for grey-box attack corrupts in this case. To solve the grey-box defence problem, we introduce a collaborative architecture into the framework.

\subsubsection{Collaborative training for grey-box attack}\label{w_version}
We develop a framework that not only defends against transferring black-box attacks but also stops generating adversarial example using the oracle, in which the adversary has a priori knowledge of both the model and the defending strategy.

We add a gradient lock unit $g$, between logits $Z=\{z_1, z_2, \ldots, z_n\}$ and logits $Z'=\{z'_1, z'_2, \ldots, z'_n\}$. The input of $g$ is the element-wise sum of $Z$ and $Z'$ (i.e., $g(Z+Z')$). We then calculate the Hadamard product of $Z+Z'$ and a random vector $\alpha$. Particularly, we define $Z*=g(Z+Z')=\alpha\odot(Z+Z')$, wherein $\alpha=\{\alpha_1, \alpha_1, \ldots, \alpha_n\}$ follows a Bernoulli distribution $Bernoulli(p)$. $\alpha$ is re-generated before every feed-forward batch. Therefore, $\alpha$ randomly resets activations to 0 in the feed forward procedure. This generates random zero gradients during the process of crafting adversarial examples. Moreover, to overcome the effect of the activation rest during the training procedure, we produce a constant gradient of $\displaystyle \mathop{\E}_{p\sim Bernoulli(p)} = p$ between $Z*$ and $Z$/ $Z'$. The parameter $p$ is known to the defender but is hidden from the attackers. In addition, attackers cannot modify or remove $g$ during calculating gradients. These constrains are reasonable since the real-world attackers also cannot alter the architecture or parameters of a victim model.

When the model is put into use, the outputs through $Z^*$ and $Z'$ will then be checked against the $Classmap$ from Section \ref{robust_ident}. If the outputs match the encoding relationship in the $Classmap$, the output is credible. Otherwise, the input example is identified as an adversarial example. Therefore, our defending method is to detect and reject adversarial examples. Furthermore, the regularised model output from $Z^*$ can complement the detection module once there is a mis-detection.

\begin{table}[t]
\caption{Model Architectures}
\label{Model_architecture}
\centering
\begin{tabular}{l l l l}
\hline
\textbf{Layer Type} & \textbf{Oracles} & \textbf{$S_{Cifar10}$} & \textbf{$S_{MNIST}$}\\
\hline
Convo+ReLU & 3$\times$3$\times$32 & 3$\times$3$\times$64 & 3$\times$3$\times$32\\
Convo+ReLU & 3$\times$3$\times$32 & 3$\times$3$\times$64 & 3$\times$3$\times$32\\
Max Pooling & 2$\times$2 & 2$\times$2 & 2$\times$2\\
Dropout & 0.2 & - & -\\ 
Convo+ReLU & 3$\times$3$\times$64 & 3$\times$3$\times$128 & 3$\times$3$\times$64\\
Convo+ReLU & 3$\times$3$\times$64 & 3$\times$3$\times$128 & 3$\times$3$\times$64\\
Max Pooling & 2$\times$2 & 2$\times$2 & 2$\times$2\\
Dropout & 0.2 & - & -\\
Convo+ReLU & 3$\times$3$\times$128 & - & -\\
Convo+ReLU & 3$\times$3$\times$128 & - & -\\
Max Pooling & 2$\times$2 & - & -\\
Dropout & 0.2 & - & -\\
Fully Connected & 512 & 256 & 200\\
Fully Connected & - & 256 & 200\\
Dropout & 0.2 & - & - \\
Softmax & 10 & 10 & 10\\
\hline
\end{tabular}
\end{table}

\begin{table}[t]
\caption{Adversarial Examples for Evaluation}
\label{Evaluation_datasets}
\centering
\begin{tabular}{c c c c c c}
\hline
\textbf{Dataset} & \textbf{$FGSM$} & \textbf{$IGS$} & \textbf{$Deepfool$} & \textbf{$C\&W L_{2}$}\\
\hline
\textbf{MNIST} & 1,000 & 1,000 & 1,000 & 1,000\\
\textbf{Cifar10} & 1,000 & 1,000 & 1,000 & 1,000\\
\hline
\end{tabular}
\end{table}

\section{Evaluation} \label{results}
In this section, we present the evaluation on our proposed method on defending against the state-of-the-art attacking methods. We first evaluated the robustness of our defence against FGSM, IGS, JSMA, and Deepfool, in black-box setting. Then we evaluated the detection performance against high-confidence C$\&$W attacks in black-box and grey-box settings.
We ran our experiments on a Windows server with CUDA supported 11GB GPU memory, Intel i7 processor, and 32G RAM. In the training of our multi-task model and the evaluation of our defence against the fast gradient based attack, we adopted our implementation of FGSM. For Carlini$\&$Wagner attack, we employed the implementation from the paper \cite{carlini2017}. For other attacks, we used the implementations provided in Foolbox \cite{rauber2017foolbox}.

\subsection{Models, data, and attacks}
We implemented one convolutional neural network architecture as an oracle. To simplify the evaluation, we used the same architecture as the oracle for both $MNIST$ and $CIFAR10$ datasets. The oracle for the datasets are denoted as $O_{MNIST}$ and $O_{CIFAR10}$, respectively. The oracle achieved $99.5\%$ accuracy on 10,000 $MNIST$ test samples, and $80.1\%$ on 10,000 $CIFAR10$ test samples. The architectures of the oracle model are depicted in Table \ref{Model_architecture}.

We first evaluated CMT in a black-box setting against five state-of-the-art attacks, namely FGSM, IGS, Deepfool, JSMA, and Carlini$\&$ Wagner $L_{2}$ (C$\&$W). Then, we evaluated our defence against the C$\&$W attack in both black-box and grey-box settings. We selected the $L_{2}$ version of the C$\&$W attack in the evaluation because it is fully differentiable. Compared to $L_0$ and $L_\infty$ attacks, it can better find the optimal adversarial examples that will lead to misclassification. In our evaluation, we set a step size of $0.3$ in FGSM for $MNIST$, and $0.1$ for $CIFAR10$, in order to transfer the attack from the substitute to the oracle. We set a step size of $0.01$ and a maximal allowed iteration of $100$ in IGS. For C$\&$W $L_{2}$ attack, we used a step length of $0.01$ and a maximal iteration of $10,000$. We set the parameter $\kappa$ to $40$, as the setting used to break a black-box distilled network in the original paper \cite{carlini2017}, where the experiment produced high-confidence adversarial examples. Later, we also evaluated the performance of our defence under different $\kappa$ values.

In the evaluation session, we crafted 1,000 successful $MNIST$ adversarial examples and 1,000 successful $CIFAR10$ adversarial examples as the adversarial test dataset, for each type of the attacks. We summarised the sizes of all adversarial datasets in Table \ref{Evaluation_datasets}. We did not evaluate the detection performance of our defence in a black-box setting, since 1). our defence leverages adversarial training to proactively defend black-box attacks; 2). the detection defence has been evaluated in a stricter grey-box setting. However, we tested the robustness gain brought by our defence against the transferring black-box attacks.

In the case of black-box attacks, we trained two substitute models, which are named as $S_{MNIST}$ and $S_{CIFAR10}$, to search for adversarial examples. The architecture of the substitutes is summarised in Table \ref{Model_architecture}. The trained substitutes achieved an equivalent performance with the models used in the previous papers \cite{papernot2016distillation, carlini2017, meng2017magnet}. $S_{MNIST}$ achieved $99.4\%$ classification accuracy on 10,000 testing $MNIST$ examples, while $S_{CIFAR10}$ achieved $78.6\%$ accuracy on 10,000 $CIFAR10$ test samples. 

\begin{table*}[t]
\caption{Classification Accuracy on Non-targeted Black-box Adversarial Examples}
\label{Black_Accuracy}
\centering
\begin{tabular}{c|c|c c c c c c}
\hline
\textbf{Model} & & \textbf{$FGSM$} & \textbf{$IGS$} & \textbf{$Deepfool$} & \textbf{$JSMA$} &  \textbf{$C\&W L_{2}$}\\
\hline
\multirow{3}{*}\textbf{$O_{MNIST}$} 
& \textbf{$L_2$ distortions} & 5.69 & 5.13 & 4.65 & 4.11 & 32.05\\
& \textbf{Original} & 0\% & 0\% & 0\% & 0\% & 0\%\\
& \textbf{Adv} & 82.7\% & \textbf{81.3\%} & 88.1\% & \textbf{85.3\%} & 0.8\%\\
& \textbf{CMT} & \textbf{84.2\%} & 79.5\% & \textbf{89.3\%} & 81.3\% & \textbf{1.2\%}\\
\hline
\multirow{3}{*}\textbf{$O_{CIFAR10}$}
& \textbf{$L_2$ distortions} & 5.42 & 4.19 & 3.97 & 3.69 & 30.18\\
& \textbf{Original} & 0\% & 0\% & 0\% & 0\% & 0\%\\
& \textbf{Adv} & 78.8\% & 75.2\% & \textbf{86.7\%} & 77.1\% & 3.1\%\\
& \textbf{CMT} & \textbf{81.8\%} & \textbf{80.2\%} & 83.7\% & \textbf{79.3\%} & \textbf{4.5\%}\\
\hline
\hline 
\end{tabular}
\end{table*}

\subsection{Defending low-confidence adversarial examples}\label{Black_box_eval}
We present the performance of CMT against low-confidence black-box adversarial examples in this section. First, given $S_{MNIST}$ and $S_{CIFAR10}$, we used the aforementioned four attacks to craft adversarial sets whose sizes are listed in Table \ref{Evaluation_datasets}. Then, we attacked the oracles by feeding the adversarial examples into $O_{MNIST}$ and $O_{CIFAR10}$, respectively. We adopted a non-targeted version of each of the above attacks since the non-targeted adversarial examples perform better in terms of transferring between models.

Robustness towards adversarial examples is a critical criterion to be assessed for the protected model. For a black-box attack, we measured the robustness by investigating the performance of our defence on tackling typical low-confidence black-box adversarial examples, which locate in the vicinity of model decision boundaries. To demonstrate the effectiveness of our method, we compared CMT defence with standard adversarial training (Adv) \cite{Goodfellow2014Explaining}. We fed the $10,000$ adversarial examples into the protected $O_{MNIST}$ and $O_{CIFAR10}$, and then we checked the classification accuracy of the label output through $Z^*$. The results of the classification accuracy are listed in Table \ref{Black_Accuracy}. It shows that our defence is on par with the standard adversarial training.

It can be found that, CMT improved the classification accuracy of the oracle in most of the cases, except for C$\&$W attacks. The reason is that the C$\&$W attacks successfully bypass the black-box defence because the confidence of the generated example is set to a high value (i.e. $\kappa=40$). The nature of the black-box defence is to regularise the position of the decision boundaries of the models, such that adversarial examples near the decision boundary become invalid. However, the defence can be easily bypassed by adversarial examples that are perturbed with larger budget to produce higher adversarial classification confidences. This vulnerability also suggests that we need a more effective defence for C$\&$W attacks.

\subsection{Defending high-confidence adversarial examples}
We evaluated CMT against high-confidence adversarial examples crafted by C$\&$W attacks in this section. The attacking confidence in C$\&$W attack was changeable through adjusting the hyper-parameter $\kappa$ in the adversarial objective function. Large $\kappa$ value leads to producing high-confidence adversarial examples. Therefore, C$\&$W attack can even achieve remarkable attacking performance through transferring attack in black-box settings. In this case, our defence relied on detection mechanism to mitigate high-confidence C$\&$W attacks. Herein, we evaluated the performance of detecting black-box C$\&$W examples crafted using different $\kappa$ values.

\subsubsection{Defending transferring C$\&$W adversarial examples}
\begin{figure}[t]
\center
\includegraphics[width=0.8\linewidth]{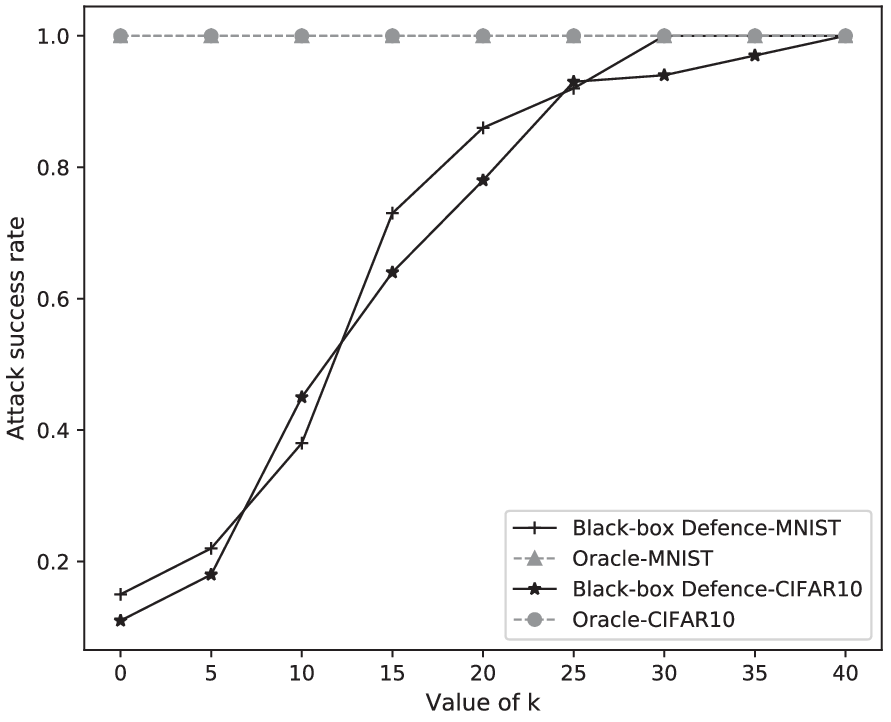}
\caption{The rate of successful transferred adversarial examples by C$\&$W attack, in black-box setting. Our black-box defence decreased the success attack rate when $\kappa$ is under 10. However, when $\kappa$ became higher, the black-box defence turned out to be invalid.}
\label{Black_box_transfer}
\end{figure}

In a black-box setting, the high-confidence C$\&$W examples can better transfer from a substitute to an oracle model. We first measured the robustness gain of CMT towards C$\&$W examples. To demonstrate the robustness gain, we measured the success rates of the transferring attacks which changed the output results from the $Z^*$. The successful transfer rate from the substitutes to the oracles are plotted in Fig.\ref{Black_box_transfer}. When $\kappa$ was set to a high value, the adversarial examples could still successfully led to misclassification of the oracles, since the nature of our black-box defence is similar to the adversarial training. That is, it regularises the decision boundaries of the oracles to invalidate the adversarial examples near the decision boundaries. But unfortunately, the high-confidence examples are usually far away from the decision boundaries. Therefore, a detection-based defence should be introduced to mitigate the high-confidence adversarial examples. Nevertheless, compared to the unprotected oracles. our proactive defence kept the attacking success rate below $40\%$ when $\kappa$ was less than $10$.

\subsubsection{Detecting high-confidence adversarial examples}
\begin{figure}[t]
\center
\includegraphics[width=0.8\linewidth]{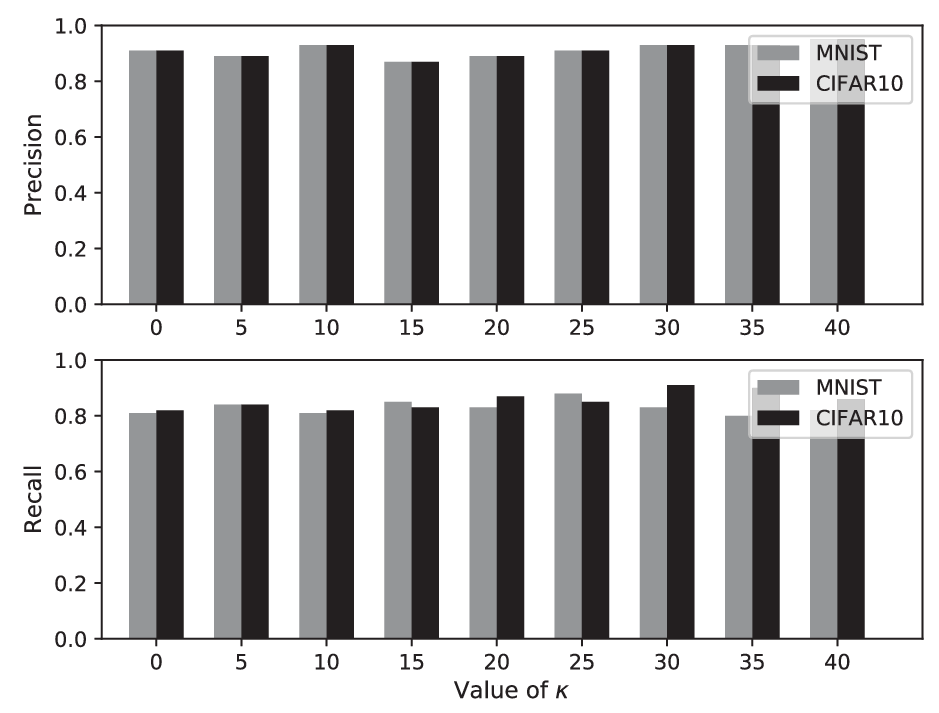}
\caption{The precision and recall of detecting C$\&$W adversarial examples in the black-box setting.}
\label{CW_Black_PR}
\end{figure}

In this section, we evaluated the performance of CMT on detecting high-confidence C$\&$W attacks. For each $\kappa$ value, we crafted 1,000 adversarial examples given the substitutes $S_{MNIST}$ and $S_{CIFAR10}$. We then mixed 1,000 benign examples into each group of adversarial examples to form evaluation datasets. We measured the precision and the recall of CMT on detecting the adversarial examples crafted under varying $\kappa$ values. The precision is calculated as the percentage of the genuine adversarial examples in the examples detected as adversarial. The recall is the percentage of adversarial examples detected from the set of adversarial examples. The precision values and the recall values are plotted in Fig.\ref{CW_Black_PR}. The precision and recall values increased with the confidence of the attacks. CMT achieved above $80\%$ detection precision on $MNIST$ and $CIFAR10$ adversarial examples when the attack confidence rose beyond 30. At the same time, the detection recall went above $70\%$ on adversarial examples that had a confidence larger than 30.

Next, we compared our defence with two state-of-the-art defences, namely Local Intrinsic Dimensionality (LID) \cite{ma2018characterizing} and DeepFense \cite{rouhani2018deepfense}, in terms of detection performance and defence capability. We first verified the capability of the defences towards black-box and grey-box attacks. The comparison is shown in Table \ref{AUC_Comparison}. Second, we compared Area Under Curve (AUC) scores of the defences on detecting high-confidence C$\&$W attacks with a $\kappa$ value of $40$. The AUC score is the area under a Receiver Operating Characteristic (ROC) curve which measures True Positive (TP) rates against False Positive (FP) rates under different detection thresholds. Herein, an AUC score of 0.5 indicates a random decision while an AUC of $1$ implies an ideal model. To obtain the AUC scores, we trained a Logistic Regression (LR) model on-the-fly during training CMT, to separate benign examples and adversarial examples. The LR detector outputs binary detection results given the softmaxed $Z'$ as inputs. During the comparison, we selected the recommended hyper-parameters for LID and DeepFense in our comparison. Specifically, we compared with DeepFense that uses $4$ latent defenders. As for LID, we adopted $20$ as the number of nearest neighbours for calculating the LID. We selected $3.79$ and $0.26$ as the bandwidths for $MNIST$ and $CIFAR10$, respectively. The comparison results are shown in Table \ref{AUC_Comparison}. Based on the comparison, CMT outperformed DeepFense and LID on $CIFAR10$ examples, and achieved comparable performance on $MNIST$ dataset. Moreover, CMT integrates both proactive defence (\textit{i.e.}, defence that invalidates attacks by improving the model robustness) and reactive defence (\textit{i.e.}, defence that detects adversarial examples), while LID and DeepFense are both reactive defences. Instead of detecting attacks and then auditing model outputs, the integration ensures that borderline attacks can be removed before the attacks take effect.

\begin{table*}
\caption{Comparison of AUC scores of CMT, LID, and DeepFense}
\label{AUC_Comparison}
\centering
\begin{tabular}{c|c c c c c c}
\hline
\multirow{2}{*}{Defence} &  \multirow{2}{*}{Black-box} & \multirow{2}{*}{Grey-box} & \multirow{2}{*}{Proactive defence} & \multirow{2}{*}{Reactive defence} & \multicolumn{2}{c}{Detection $AUC@40$}\\
& & & & & \textbf{MNIST} & \textbf{CIFAR10}\\
\hline
CMT & \checkmark & \checkmark & \checkmark & \checkmark & 0.971 & \textbf{0.947}\\						
LID & \checkmark & \checkmark & \xmark & \checkmark & 0.976 & 0.901\\
DeepFense & \checkmark & \checkmark & \xmark & \checkmark & \textbf{0.984} & 0.926\\
\hline
\end{tabular}
\end{table*}

\subsection{Tackling grey-box adversarial example generation}
We evaluated the performance of our defence against grey-box attacks. We again employed the C$\&$W attack in this section since the C$\&$W attack can produce competitive attacking results, and it is flexible for searching adversarial examples of varying attacking confidences, due to its tunable $\kappa$ hyper-parameter. The adversarial examples used in the evaluation were crafted based on the linearly-combined adversarial loss functions mentioned in Section \ref{break_g_defence}. To measure the grey-box defence, we swept the value of $\kappa$ from 0 to 40 with step sizes of 5, and then we examined the rate of successful adversarial image generation given 100 $MNIST$ images under each $\kappa$ value. We recorded the successful generation rates of targeted and non-targeted $MNIST$ adversarial examples based on the defended oracle. For targeted attacks, we randomly set the target labels during the generation process. The generation rates are recorded in Table.\ref{Generation_Rate_CW}.

\begin{table*}[t]
\caption{The Successful Generation Rate of C$\&$W Attack in grey-box Setting}
\label{Generation_Rate_CW}
\centering
\begin{tabular}{c|c|c c c c c c c c c}
\hline
\textbf{Attack} & \textbf{Model} & $\kappa=0$ & $5$ & $10$ & $15$ & $20$ & $25$ & $30$ & $35$ & $40$\\
\hline
\multirow{2}{*}\textbf{Non-targeted} & \textbf{Original} & $100\%$ & $100\%$ & $100\%$ & $100\%$ & $100\%$ & $100\%$ & $100\%$ & $100\%$ & $100\%$\\
& \textbf{Defended} & $8\%$ & $0\%$ & $0\%$ & $0\%$ & $0\%$ & $0\%$ & $0\%$ & $0\%$ & $0\%$\\
\hline
\multirow{2}{*}\textbf{Targeted} & \textbf{Original} & $100\%$ & $100\%$ & $100\%$ & $100\%$ & $100\%$ & $98\%$ & $96\%$ & $92\%$ & $90\%$\\
& \textbf{Defended} & $0\%$ & $0\%$ & $0\%$ & $0\%$ & $0\%$ & $0\%$ & $0\%$ & $0\%$ & $0\%$\\
\hline
\end{tabular}
\end{table*}

\begin{figure}[t]
\center
\includegraphics[width=0.8\linewidth]{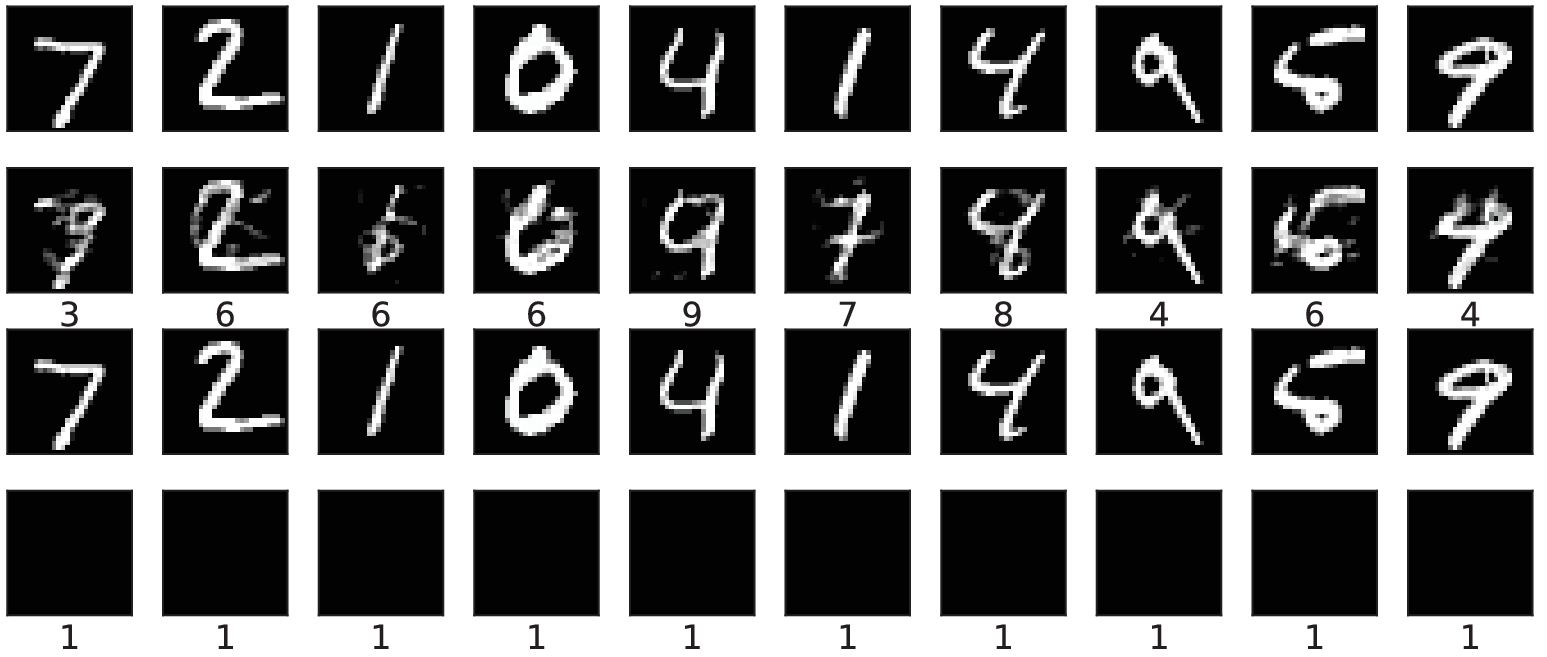}
\caption{The adversarial $MNIST$ images generated using non-targeted C$\&$W attack when $\kappa=40$. The first row and the third row are the original images. The second row is the generated adversarial image based on the original model (classification results are as the labels below the second row). The fourth row is the failed generation after applying our defence on the model.}
\label{Non_targeted_40k}
\end{figure}

It can be found that the rates of finding valid adversarial examples are kept at a low level, especially when the values of $k$ are high. We include some of the C$\&$W adversarial examples generated with/without our defence in Fig.\ref{Non_targeted_40k}.

\subsection{Trade-off on benign examples}
In this section, we evaluated the trade-off on normal example classification and the FP rate of detection when the input examples are benign. We also measured the overhead added on the oracles by CMT.

First, we evaluated the accuracy of the classification results outputted through $Z^*$, after our protections were applied on the oracle. We used both $CIFAR10$ dataset and $MNIST$ datasets in the evaluation. For each dataset, we drew $10,000$ benign examples and fed them into the defended oracles. The results are shown in Table \ref{Benign_Classification}. Herein, the classification accuracy was measured by the accuracy of the output classification through $Z$. It could be found that, for $O_{MNIST}$, our defence had no decrease in the classification accuracy. On $CIFAR10$ task, our defence decreased the accuracy by $2.1\%$. The trade-off was within the acceptable range, considering the improvements in defending adversarial examples. The runtime for detecting the 10,000 benign examples from $MNIST$ and $CIFAR10$ was also recorded in Table \ref{Benign_Classification}. CMT added an extra overhead of $2.6\times10^{-4}$s to the protected models for detecting an $MNIST$ example. An overhead of $4.81\times10^{-4}$s per example was added when detecting $CIFAR10$ examples. The throughput of the protected models was still kept at a reasonable level. On the other hand, the defence only increased $40\sim50$MB of GPU memory usage.

Next, we assessed the mis-detection rate of our defence. We fed 10,000 benign $MNIST$ examples and 10,000 $CIFAR10$ benign examples into the corresponding defended oracles to check how many of the examples were incorrectly recognised as adversarial. As the results suggest, CMT achieved $0.09\%$ mis-detection rate on $MNIST$ dataset. For $CIFAR10$ dataset, the mis-detection rate was $3.92\%$. According to the results, CMT could accurately separate the adversarial examples from the benign examples.

\begin{table}
\caption{The classification accuracy and the runtime on 10,000 benign examples}
\label{Benign_Classification}
\centering
\begin{tabular}{c|c c c}
\hline
		& Model & Without CMT & With CMT\\
\hline
\multirow{2}{*}\textbf{$O_{MNIST}$} & Accuracy & 99.4\% & 99.4\%\\
							  & Runtime & 0.61s & 3.21s\\
							  & Total GPU usage & 8855MB & 8899MB\\
\multirow{2}{*}\textbf{$O_{CIFAR10}$} & Accuracy & 78.6\% & 76.5\%\\
							   & Runtime & 1.53s & 6.34s\\
							   & Total GPU usage & 8903MB & 8961MB\\
\hline
\end{tabular}
\end{table}

\section{Justification of the defence}\label{justification}
In this section, we present the justification on the mechanism of our defence on both black-box and C$\&$W attacks.

\subsection{Classmap evaluation}\label{cmap}
\begin{figure*}[t]
\center
\includegraphics[width=0.8\linewidth]{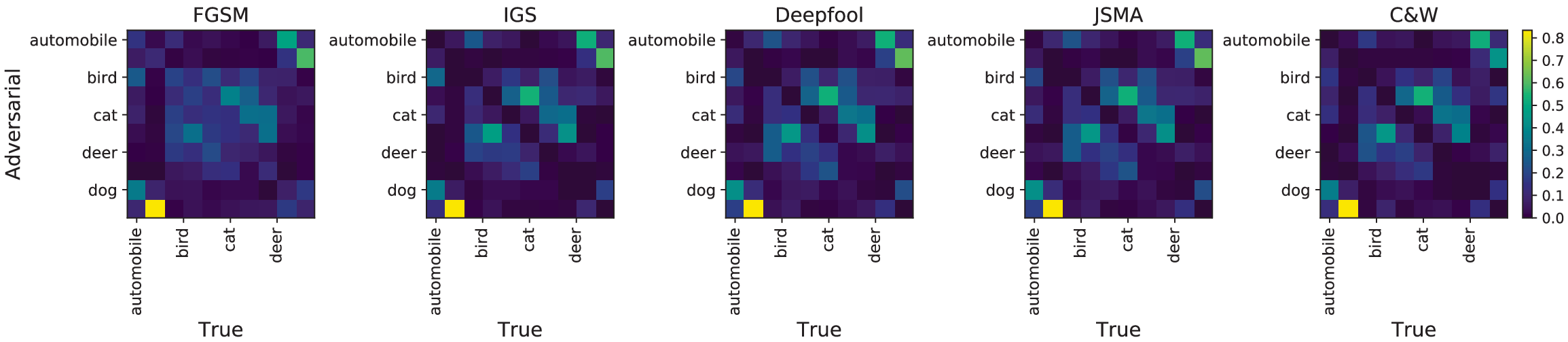}
\caption{The misclassification $Classmaps$ of FGSM, IGS, Deepfool, JSMA, and C$\&$W examples from low-confidence attacks. It can be found that all the classmaps share a similar pattern.}
\label{Classmap_eval}
\end{figure*}

We evaluated the generalisation of the $Classmap$ extracted from different attacking examples based on $CIFAR10$ data. First, we generated $Classmaps$ from low-confidence FGSM, IGS, Deepfool, JSMA, and C$\&$W attacks. Given a trained model, each attack generated 1,000 borderline adversarial examples from the training data. We set $\kappa=1$ in the C$\&$W attacks. We set the learning rate to be $0.01$ and the maximal iteration to be $100$ in the IGS attack. The learning rate of FGSM is set to $0.1$. The distributions of the misclassification are displayed in Fig.\ref{Classmap_eval}. All the $Classmaps$ shared a similar pattern. This means that the $Classmaps$ learnt from the first-order FGSM examples can be generalised to other types of adversarial examples when the attacks search adversarial examples in the vicinity of a decision boundary. The generalisation of the $Classmap$ ensured that the detector trained on an attack could generalise across different attacks. In other words, the defence is not required to know the details of an attack before defending it. As an example, herein, we used FGSM to generate a $Classmap$ for defending C$\&$W attacks.

\begin{figure}[t]
\center
\includegraphics[width=1\linewidth]{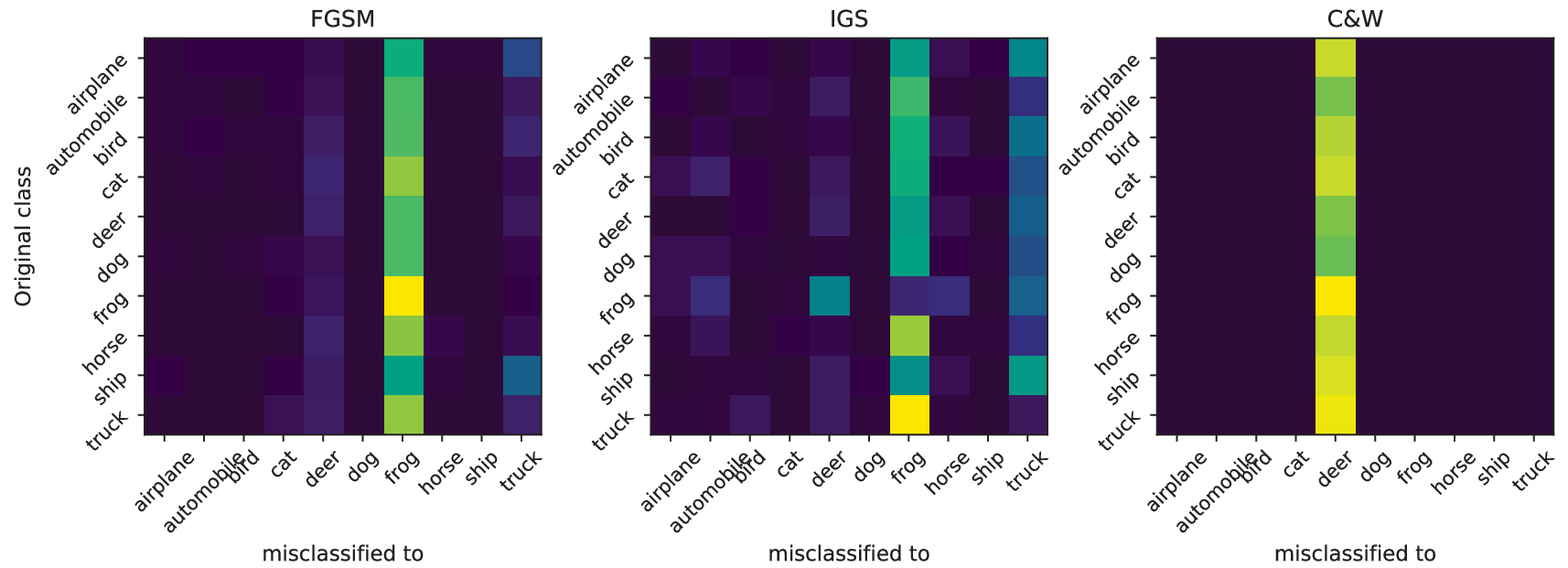}
\caption{The misclassification class map of FGSM, IGS and C$\&$W examples from high confidence attack. It can be found that the classmaps diverges from the classmaps of low-confidence attack.}
\label{Classmap_eval_2}
\end{figure}

Next, we generated $Classmaps$ from high-confidence FGSM, IGS, and C$\&$W attacks. We crafted $1,000$ examples of the training data for each attack. We set a $\kappa=40$ in the C$\&$W attacks. We set a learning rate of $0.1$ and a maximal iteration of $100$ in the IGS attacks. The learning rates of the FGSM were set to $0.5$. The $Classmaps$ of high-confidence attacks is then plotted in Fig.\ref{Classmap_eval_2}. It can be observed that the $Classmap$ differs from that of low-confidence attacks. This difference enables our defence to detect high-confidence attacks.

\subsection{Defending low-confidence black-box attack}
For a black-box attack, the adversarial training introduced in this model can effectively regularises the decision boundaries of the model to tackle adversarial examples near the decision boundaries. Compared to vanilla adversarial training \cite{Szegedy2013Intriguing}, our method can further increase the distance required for moving adversarial examples. Crafting adversarial examples based on deep neural networks leverages back-propagation of objective loss to adjust the input pixel values. Hence, the adversary is actually moving the data point in the feature space according to the gradient direction $\bigtriangledown_{x_{adv}}{L(x_{adv}, y)}$ to maximise the adversarial loss $L$ (a non-targeted attack in a black-box setting). Suppose an adversarial example is found after $n$ steps of gradient decent, for each step of gradient decent, we have a step size $\alpha$, the total perturbation can be approximately calculated as:

\begin{equation}
\begin{split}
\delta & =x_{n}-x_{0}=\alpha\cdot(\bigtriangledown_{x_{n-1}}{L^{n-1}}+\bigtriangledown_{x_{n-2}}{L^{n-2}}\\&+\ldots+\bigtriangledown_{x_{1}}{L^{1}}+\bigtriangledown_{x_{0}}{L^{0}})
\end{split}
\end{equation}

According to Section \ref{robust_ident}, the adversary relies on the gradient decent based updates to gradually perturb image until it becomes adversarial. An adversarial example is usually found within a $\delta$-bounded hypersphere centred by the original example. A low-confidence example can be found with smaller $\delta$. Based on the discussion in Section \ref{vds}, the adversarial examples in the robust class $l_{r}$ with respect to the original example. For non-targeted attacks, given a small $\delta$ and an example whose original label is $l$, it is likely to find an adversarial example being classified into a class other than $l_{r}$. Similarly, in the targeted attacks, $l_{r}$ is more difficult to be used as the adversarial target label. Suppose the total efforts for maximising/minimising $L(x:F(x)=l,l_{i}:i\neq r)$ is $\delta_i$, the total efforts for maximising/minimising $L(x:F(x)=l,l_{r})$ is $\delta_r$,we have $\delta_r > \delta_i$. When the training objective includes a term that contains the robust label $y_{robust}$, the output of the trained model could be treated as a linear combination of the outputs trained from $l$ and $l_{r}$. Therefore, the required total efforts for changing the combined output becomes higher.

From the perspective of a classifier decision boundary, our multi-task training method has also increased the robustness of the model against black-box examples. The robust label regularisation term actually moves the decision boundary towards the robust class. Compared to the traditional adversarial training, which tunes the decision boundary depending merely on the generated adversarial data points, our regularisation further enhances the robustness of the model towards nearby adversarial examples.

\subsection{Detecting C$\&$W attacks}
We provide a brief analysis on why our method can defend C$\&$W attacks here. A C$\&$W attack mainly relies on the modified objective function to search for adversarial examples, given the logits from the model. By observing the objective function $f(x_{adv})=max(max\{Z(x_{adv})_{i}:i\ne l\}-Z(x_{adv})_{l},-\kappa)$, we can find out that the objective is actually to increase the value of the logits corresponding to the desired $l$ class, until the difference between $l$ class and the second-to-the-largest class reaches the upper bound defined by $\kappa$. The optimisation process can be interpreted as adjusting the input pixels along the direction of the gradient that maximises the logits difference.

In a black-box setting, when we adopt the collaborative multi-task training as a defence, the model actually modifies the output logits to have high outputs not only on the position corresponding to the ground truth class, but also on the position corresponding to the robust class of the current ground truth. In the black-box setting, the defence is hidden from the attacker. The attacker crafts adversarial examples solely based on the oracle model without the robust logits branch. Hence, the adversarial objective function is not a linear combination of $L(Z(x),t)$ and $L(Z(x),t')$, but a single loss $L(Z(x),t)$. The crafted adversarial example can only modify the output from $Z$ to the adversarial target $t$, but the output through $Z'$ is not guaranteed  to be the corresponding robust label of $t$ in the $Classmap$.

Based on the evaluation on the $Classmap$, the misclassification patter of high-confidence C$\&$W examples is different from that of low-confidence examples. In other words, give a low-confidence adversarial example and a high-confidence example in a same class, their outputs through $Z'$ are mostly different. Therefore, high-confidence examples that bypass the black-box defence can be detected.

\subsection{Thwarting adversarial example generation}
\begin{figure}[t]
\center
\includegraphics[width=0.8\linewidth]{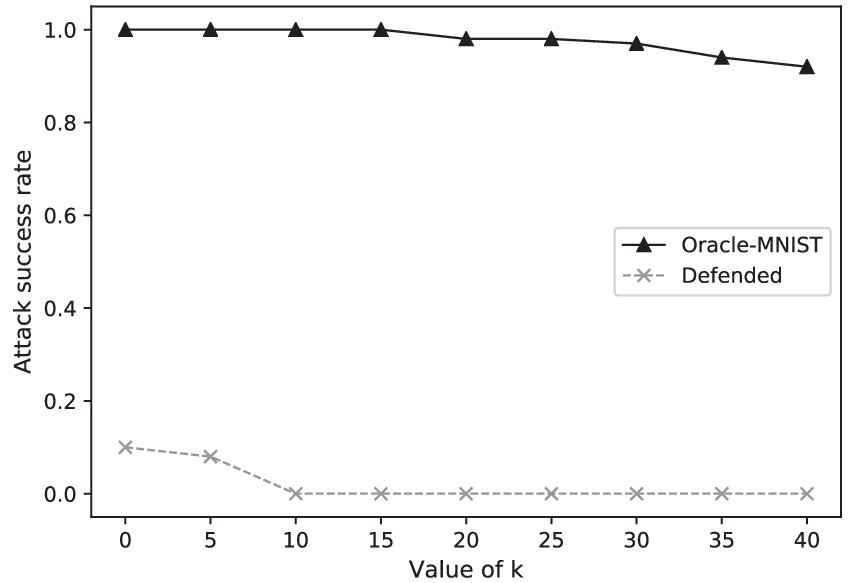}
\caption{The rate of successfully generated C$\&$W examples under different confidences. It can be observed that our defence effectively prevents C$\&$W from generating successful adversarial examples.}
\label{cw_gen}
\end{figure}

The added defence stops attackers from generating adversarial examples. In this section, we provide an analysis on how our defence proves the C$\&$W attack from generating adversarial examples. To study the defence gain brought by the gradient lock proposed in the paper, we apply the gradient lock after the logits of $O_{MNIST}$ and train the protected $O^p_{MNIST}$. We train $O^p_{MNIST}$ by setting $p=0.3$. Then, given 1,000 $MNIST$ examples, we try to generate non-targeted C$\&$W examples based on $O^p_{MNIST}$ and $O_{MNIST}$, separately. We record the successful generation rates of C$\&$W examples with different confidence values. The success rates of C$\&$W attack under different confidence are plotted in Fig.\ref{cw_gen}. It can be observed that our defence effectively reduces the rate of generating successful C$\&$W examples.

\section{Related work} \label{related_work}
\subsection{Attacking methods}
Based on the method of generating adversarial example, current attacking methods can be divided into the optimisation based attacks and the forward derivative based attacks. The optimisation based attack sets an adversarial objective function and optimises the example feature to achieve the optimum. FGSM uses a single gradient descent step to slightly perturb the input example \cite{Goodfellow2014Explaining}. Subsequently, an iterative gradient sign based method was proposed in \cite{kurakin2016adversarial}. L-BFGS based attack optimises a box-constrained adversarial objective to find adversarial example \cite{Szegedy2013Intriguing}. Furthermore, DeepFool iteratively finds the minimum perturbations of images \cite{Moosavidezfooli2016DeepFool}. Last but not the least, the Carlini$\&$Wagner attack is proposed to optimise a specially designed adversarial loss to craft adversarial example with changeable attacking confidence \cite{carlini2017}. Otherwise, the forward derivative based attacks generate perturbations based on the Jacobian of model outputs with respect to input features. The most representative attack in this category is JSMA \cite{Papernot2016The}. 

To make adversarial examples more imperceptible for human beings, there are methods using different distortion metrics. For example, an image can be perturbed in HSV color space to generate adversarial example \cite{hosseini2018semantic}. Additionaly, an image can be rotated to be adversarial \cite{engstrom2017rotation}. Beyond the mentioned attacks, there are attacks modified towards different systems. Adversarial examples have been designed towards applications such as object detection and segmentation \cite{xie2017adversarial, brown2017adversarial, Athalye2017, lu2017adversarial}, reading comprehension \cite{Jia2017}, text classification \cite{Ebrahimi2018}, and malware detection \cite{Grosse2017a}.

\subsection{Defensive methods}
Defensive methods can be broadly categorised into proactive defences and reactive defences. In the category of proactive defence, a model is trained to be robust towards adversarial examples. For example, the adversarial training techniques incorporates adversarial gradients in the training phase to enhance model robustness \cite{Szegedy2013Intriguing,lyu2015unified,tramer2017ensemble, madry2017towards}. Other defences rely on Lipschitz condition to make DNNs insensitive to adversarial perturbations \cite{Gu2015Towards}. Similarly, the defensive distillation is used to reduce the sensitivity of DNNs towards adversarial perturbations \cite{papernot2016distillation}. \cite{cisse2017parseval}.

For most of the reactive defence methods, a second model is adopted to detect examples with adversarial perturbation. For instance, Metzen et al. attached detectors on a model and trained it with adversarial examples \cite{metzen2017detecting}. Another method employs support vector machine to classify the output from the high-level neural network layer \cite{lu2017safetynet}.  A statistical method is proposed to detect adversarial example batches \cite{grosse2017statistical}. Lately, there is a detection method that detects adversarial examples based on the local intrinsic dimensionality of layer representations \cite{ma2018characterizing}. An online accelerated defence relies on multiple redundant models to validate input examples and then detect adversarial examples \cite{rouhani2018deepfense}. For other detection methods, they align with the idea of observing the changes of an example after applying a certain transformation on it \cite{song2017pixeldefend, xu2017feature, meng2017magnet}. For example, MagNet relies on autoencoder to reconstruct adversarial examples to normal example, and detect adversarial example based on the reconstruction error and output probability divergence \cite{meng2017magnet}.

\section{Discussion}\label{discussion}
Current attacks and defences are largely considered as threat-model-dependent. As the attacking methods based on gradients, except C$\&$W attack, are unable to attack distilled network in a grey-box setting, a black-box attack that adopts these methods is more practical and harmful. This is why we evaluated the performance of our defence in a black-box setting. As a special case, the C$\&$W attack claims that it can break the defensive distillation in the white-box setting since it searches for adversarial examples based on the logits instead of the softmax outputs. Hence, the C$\&$W attack can bypass the vanishing gradient mechanism introduced by defensive distillation on the softmax layer. However, it is a very strong assumption to have access to the logits itself, which actually defines a white-box attack. However, a substitute can be used together with high confidence C$\&$W attack to bypass the distilled network in a black-box setting.

Scaling to adversarial examples from large datasets such as ImageNet is a challenging task for defending methods, including CMT. Complex data manifold and high dimensionality of the large datasets hinder the defence from characterising adversarial examples. For example, it is difficult to find ImageNet adversarial examples by first-order attacks, such as FGSM, within a plausible distortion budget. Reversely, successful adversarial examples of large datasets have more non-linearity and more complexity than that of small datasets, which poses a challenge to categorising the adversarial examples. Generally, the models trained by large datasets are more robust than that trained by simple datasets~\cite{madry2017towards}. Currently, the scalable defending methods mainly rely on input transformations to eliminate the effect of adversarial perturbations (\textit{e.g.,}~\cite{guo2017countering, gupta2019ciidefence}). Therefore, the robustness of large models actually encourages the defence. So far, the scalability is still an open issue in many other state-of-the-art methods (\textit{e.g.,}~\cite{buckman2018thermometer, madry2017towards, zhang2019defense}).

CMT can be further improved by incorporating randomness into the defence architecture. Second, the attacks that employ forward derivatives (\textit{e.g.} JSMA \cite{papernot2016limitations}) in the grey-box setting can still effectively find adversarial examples. This is because our defence essentially tackles gradient-based adversarial example searching. However, our defence is still functional towards black-box JSMA examples due to the regularised training process. At last, our grey-box defence is based on the gradient masking. This can be improved in our future work.

\section{Conclusion and Future Work}\label{conclusion}
In this paper, we proposed a novel defence against black-box attacks and grey-box attacks on DNNs. According to the evaluation, our defence can defend both black-box attacks and grey-box attacks without knowing the details of the attacks in advance. As for the shortcomings of our approach, first, the quality of the $Classmap$ will directly affect the performance of the detection rate for adversarial examples. We use a large volume of non-targeted adversarial examples to estimate the encoding rules among class labels. However, the quality of the estimation is affected by the employed attacking method. Second, our defence is proposed based on the properties of non-targeted attacks and is designed for the non-targeted attacks. For targeted attacks, our defence can be further strengthened by training the classmaps based on the targeted attacks. We will improve our defence in future work.

Current attacks and defences have not yet been exclusively applied to the real-world systems built on DNN. Previous studies have made attempts to attack online deep learning service providers, such as Clarifi \cite{liu2016delving}, Amazon Machine Learning, MetaMind, and Google cloud prediction API \cite{papernot2017practical}. However, there is no reported instance of attacking classifier embedded inside complex systems, such as Nvidia Drive PX2. Successful attacks on those systems might require much more sophisticated pipeline of exploiting vulnerabilities in system protocols, acquiring data stream, and crafting/injecting adversarial examples. However, once the pipeline is built, the potential damage it can deal with would be fatal. This could be another direction for future work.

\bibliographystyle{abbrv}
\bibliography{./advref}

\begin{thebibliography}{10}

\bibitem{obfuscated-gradients}
A.~Athalye, N.~Carlini, and D.~Wagner.
\newblock Obfuscated gradients give a false sense of security: Circumventing
  defenses to adversarial examples.
\newblock In {\em Proceedings of the 35th International Conference on Machine
  Learning, {ICML} 2018}, July 2018.

\bibitem{Athalye2017}
A.~Athalye and I.~Sutskever.
\newblock Synthesizing robust adversarial examples.
\newblock {\em arXiv preprint arXiv:1707.07397}, 2017.

\bibitem{brown2017adversarial}
T.~B. Brown, D.~Man{\'e}, A.~Roy, M.~Abadi, and J.~Gilmer.
\newblock Adversarial patch.
\newblock {\em arXiv preprint arXiv:1712.09665}, 2017.

\bibitem{buckman2018thermometer}
J.~Buckman, A.~Roy, C.~Raffel, and I.~Goodfellow.
\newblock Thermometer encoding: One hot way to resist adversarial examples.
\newblock {\em International Conference on Learning Representations}, 2018.

\bibitem{carlini2017adversarial}
N.~Carlini and D.~Wagner.
\newblock Adversarial examples are not easily detected: Bypassing ten detection
  methods.
\newblock {\em arXiv preprint arXiv:1705.07263}, 2017.

\bibitem{carlini2017magnet}
N.~Carlini and D.~Wagner.
\newblock Magnet and" efficient defenses against adversarial attacks" are not
  robust to adversarial examples.
\newblock {\em arXiv preprint arXiv:1711.08478}, 2017.

\bibitem{carlini2017}
N.~Carlini and D.~Wagner.
\newblock Towards evaluating the robustness of neural networks.
\newblock In {\em Proceedings of the IEEE Symposium on Security and Privacy
  (S\&P)}, pages 39--57. IEEE, 2017.

\bibitem{cisse2017parseval}
M.~Cisse, P.~Bojanowski, E.~Grave, Y.~Dauphin, and N.~Usunier.
\newblock Parseval networks: Improving robustness to adversarial examples.
\newblock In {\em International Conference on Machine Learning}, pages
  854--863, 2017.

\bibitem{Ebrahimi2018}
J.~Ebrahimi, A.~Rao, D.~Lowd, and D.~Dou.
\newblock Hotflip: White-box adversarial examples for text classification.
\newblock In {\em Proceedings of the 56th Annual Meeting of the Association for
  Computational Linguistics}, pages 31--36, 2018.

\bibitem{engstrom2017rotation}
L.~Engstrom, D.~Tsipras, L.~Schmidt, and A.~Madry.
\newblock A rotation and a translation suffice: Fooling cnns with simple
  transformations.
\newblock {\em arXiv preprint arXiv:1712.02779}, 2017.

\bibitem{evgeniou2004regularized}
T.~Evgeniou and M.~Pontil.
\newblock Regularized multi--task learning.
\newblock In {\em Proceedings of the tenth ACM SIGKDD international conference
  on Knowledge discovery and data mining}, pages 109--117. ACM, 2004.

\bibitem{Feinman2017}
R.~Feinman, R.~R. Curtin, S.~Shintre, and A.~B. Gardner.
\newblock Detecting adversarial samples from artifacts.
\newblock {\em arXiv preprint arXiv:1703.00410}, 2017.

\bibitem{girshick2016region}
R.~Girshick, J.~Donahue, T.~Darrell, and J.~Malik.
\newblock Region-based convolutional networks for accurate object detection and
  segmentation.
\newblock {\em IEEE transactions on pattern analysis and machine intelligence},
  38(1):142--158, 2016.

\bibitem{Goodfellow2014Explaining}
I.~J. Goodfellow, J.~Shlens, and C.~Szegedy.
\newblock Explaining and harnessing adversarial examples.
\newblock {\em Computer Science}, 2014.

\bibitem{grosse2017statistical}
K.~Grosse, P.~Manoharan, N.~Papernot, M.~Backes, and P.~McDaniel.
\newblock On the (statistical) detection of adversarial examples.
\newblock {\em arXiv preprint arXiv:1702.06280}, 2017.

\bibitem{Grosse2017a}
K.~Grosse, N.~Papernot, P.~Manoharan, M.~Backes, and P.~McDaniel.
\newblock Adversarial examples for malware detection.
\newblock In {\em European Symposium on Research in Computer Security}, pages
  62--79. Springer, 2017.

\bibitem{Gu2015Towards}
S.~Gu and L.~Rigazio.
\newblock Towards deep neural network architectures robust to adversarial
  examples.
\newblock {\em Computer Science}, 2015.

\bibitem{guo2017countering}
C.~Guo, M.~Rana, M.~Cisse, and L.~Van Der~Maaten.
\newblock Countering adversarial images using input transformations.
\newblock {\em International Conference on Learning Representations}, 2018.

\bibitem{gupta2019ciidefence}
P.~Gupta and E.~Rahtu.
\newblock Ciidefence: Defeating adversarial attacks by fusing class-specific
  image inpainting and image denoising.
\newblock In {\em Proceedings of the IEEE International Conference on Computer
  Vision}, pages 6708--6717, 2019.

\bibitem{he2017adversarial}
W.~He, J.~Wei, X.~Chen, N.~Carlini, and D.~Song.
\newblock Adversarial example defenses: ensembles of weak defenses are not
  strong.
\newblock In {\em Proceedings of the 11th USENIX Conference on Offensive
  Technologies}, pages 15--15. USENIX Association, 2017.

\bibitem{hosseini2018semantic}
H.~Hosseini and R.~Poovendran.
\newblock Semantic adversarial examples.
\newblock In {\em Proceedings of the IEEE conference on Computer Vision and
  Pattern Recognition (CVPR) Workshops}, pages 1614--1619. IEEE, 2018.

\bibitem{Jia2017}
R.~Jia and P.~Liang.
\newblock Adversarial examples for evaluating reading comprehension systems.
\newblock In {\em Proceedings of the 2017 Conference on Empirical Methods in
  Natural Language Processing}, pages 2021--2031, 2017.

\bibitem{kurakin2016adversarial}
A.~Kurakin, I.~Goodfellow, and S.~Bengio.
\newblock Adversarial examples in the physical world.
\newblock {\em arXiv preprint arXiv:1607.02533}, 2016.

\bibitem{li2015convolutional}
H.~Li, Z.~Lin, X.~Shen, J.~Brandt, and G.~Hua.
\newblock A convolutional neural network cascade for face detection.
\newblock In {\em Proceedings of the IEEE conference on Computer Vision and
  Pattern Recognition (CVPR)}, pages 5325--5334. IEEE, 2015.

\bibitem{liu2016delving}
Y.~Liu, X.~Chen, C.~Liu, and D.~Song.
\newblock Delving into transferable adversarial examples and black-box attacks.
\newblock {\em arXiv preprint arXiv:1611.02770}, 2016.

\bibitem{lu2017safetynet}
J.~Lu, T.~Issaranon, and D.~Forsyth.
\newblock Safetynet: Detecting and rejecting adversarial examples robustly.
\newblock {\em arXiv preprint arXiv:1704.00103}, 2017.

\bibitem{lu2017adversarial}
J.~Lu, H.~Sibai, and E.~Fabry.
\newblock Adversarial examples that fool detectors.
\newblock {\em arXiv preprint arXiv:1712.02494}, 2017.

\bibitem{lyu2015unified}
C.~Lyu, K.~Huang, and H.-N. Liang.
\newblock A unified gradient regularization family for adversarial examples.
\newblock In {\em Proceedings of the IEEE International Conference on Data
  Mining (ICDM)}, pages 301--309. IEEE, 2015.

\bibitem{ma2018characterizing}
X.~Ma, B.~Li, Y.~Wang, S.~M. Erfani, S.~Wijewickrema, M.~E. Houle,
  G.~Schoenebeck, D.~Song, and J.~Bailey.
\newblock Characterizing adversarial subspaces using local intrinsic
  dimensionality.
\newblock {\em arXiv preprint arXiv:1801.02613}, 2018.

\bibitem{madry2017towards}
A.~Madry, A.~Makelov, L.~Schmidt, D.~Tsipras, and A.~Vladu.
\newblock Towards deep learning models resistant to adversarial attacks.
\newblock {\em International Conference on Learning Representations}, 2018.

\bibitem{meng2017magnet}
D.~Meng and H.~Chen.
\newblock Magnet: a two-pronged defense against adversarial examples.
\newblock {\em arXiv preprint arXiv:1705.09064}, 2017.

\bibitem{metzen2017detecting}
J.~H. Metzen, T.~Genewein, V.~Fischer, and B.~Bischoff.
\newblock On detecting adversarial perturbations.
\newblock In {\em Proceedings of the IEEE International Conference on Learning
  Representation (ICLR)}, 2017.

\bibitem{Moosavidezfooli2016DeepFool}
S.~M. Moosavidezfooli, A.~Fawzi, and P.~Frossard.
\newblock Deepfool: A simple and accurate method to fool deep neural networks.
\newblock In {\em Proceedings of the IEEE conference on Computer Vision and
  Pattern Recognition (CVPR)}, pages 2574--2582. IEEE, 2016.

\bibitem{papernot2017practical}
N.~Papernot, P.~McDaniel, I.~Goodfellow, S.~Jha, Z.~B. Celik, and A.~Swami.
\newblock Practical black-box attacks against machine learning.
\newblock In {\em Proceedings of the ACM on Asia Conference on Computer and
  Communications Security}, pages 506--519. ACM, 2017.

\bibitem{papernot2016limitations}
N.~Papernot, P.~McDaniel, S.~Jha, M.~Fredrikson, Z.~B. Celik, and A.~Swami.
\newblock The limitations of deep learning in adversarial settings.
\newblock In {\em Proceedings of the IEEE European Symposium on Security and
  Privacy (EuroS\&P)}, pages 372--387. IEEE, 2016.

\bibitem{Papernot2016The}
N.~Papernot, P.~Mcdaniel, S.~Jha, M.~Fredrikson, Z.~B. Celik, and A.~Swami.
\newblock The limitations of deep learning in adversarial settings.
\newblock pages 372--387, 2016.

\bibitem{papernot2016distillation}
N.~Papernot, P.~McDaniel, X.~Wu, S.~Jha, and A.~Swami.
\newblock Distillation as a defense to adversarial perturbations against deep
  neural networks.
\newblock In {\em Proceedings of the IEEE Symposium on Security and Privacy
  (S\&P)}, pages 582--597. IEEE, 2016.

\bibitem{rauber2017foolbox}
J.~Rauber, W.~Brendel, and M.~Bethge.
\newblock Foolbox v0. 8.0: A python toolbox to benchmark the robustness of
  machine learning models.
\newblock {\em arXiv preprint arXiv:1707.04131}, 2017.

\bibitem{rouhani2018deepfense}
B.~D. Rouhani, M.~Samragh, M.~Javaheripi, T.~Javidi, and F.~Koushanfar.
\newblock Deepfense: Online accelerated defense against adversarial deep
  learning.
\newblock In {\em 2018 IEEE/ACM International Conference on Computer-Aided
  Design (ICCAD)}, pages 1--8. IEEE, 2018.

\bibitem{shaham2018understanding}
U.~Shaham, Y.~Yamada, and S.~Negahban.
\newblock Understanding adversarial training: Increasing local stability of
  supervised models through robust optimization.
\newblock {\em Neurocomputing}, 2018.

\bibitem{song2017pixeldefend}
Y.~Song, T.~Kim, S.~Nowozin, S.~Ermon, and N.~Kushman.
\newblock Pixeldefend: Leveraging generative models to understand and defend
  against adversarial examples.
\newblock {\em arXiv preprint arXiv:1710.10766}, 2017.

\bibitem{Szegedy2013Intriguing}
C.~Szegedy, W.~Zaremba, I.~Sutskever, J.~Bruna, D.~Erhan, I.~Goodfellow, and
  R.~Fergus.
\newblock Intriguing properties of neural networks.
\newblock {\em Computer Science}, 2013.

\bibitem{tramer2017ensemble}
F.~Tram{\`e}r, A.~Kurakin, N.~Papernot, D.~Boneh, and P.~McDaniel.
\newblock Ensemble adversarial training: Attacks and defenses.
\newblock {\em arXiv preprint arXiv:1705.07204}, 2017.

\bibitem{wang2019daedalus}
D.~Wang, C.~Li, S.~Wen, S.~Nepal, and Y.~Xiang.
\newblock Daedalus: Breaking non-maximum suppression in object detection via
  adversarial examples.
\newblock {\em arXiv preprint arXiv:1902.02067}, 2019.

\bibitem{xie2017adversarial}
C.~Xie, J.~Wang, Z.~Zhang, Y.~Zhou, L.~Xie, and A.~Yuille.
\newblock Adversarial examples for semantic segmentation and object detection.
\newblock In {\em Proceedings of the IEEE International Vonference on Computer
  Vision (ICCV)}, pages 1369--1378. IEEE, 2017.

\bibitem{xu2017feature}
W.~Xu, D.~Evans, and Y.~Qi.
\newblock Feature squeezing: Detecting adversarial examples in deep neural
  networks.
\newblock {\em arXiv preprint arXiv:1704.01155}, 2017.

\bibitem{zhang2019defense}
H.~Zhang and J.~Wang.
\newblock Defense against adversarial attacks using feature scattering-based
  adversarial training.
\newblock In {\em Advances in Neural Information Processing Systems}, pages
  1829--1839, 2019.

\end{thebibliography}
\begin{IEEEbiography}[{\includegraphics[width=1in,height=1.25in,clip,keepaspectratio]{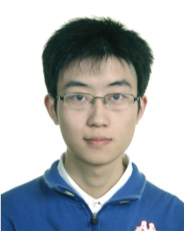}}]{Derui (Derek) Wang} received his bachelor’s degree of Engineering from Huazhong University of Science and Technology (HUST), China (2011). He then received his M.Sc. by research degree from Deakin University, Australia (2016). Currently, he is a PhD student with Swinburne University of Technology and CSIRO Data61, Australia. His research interests include adversarial machine learning, deep neural networks, applied machine learning, decision making systems, and complex networks.
\end{IEEEbiography}

\vskip 0pt plus -1fil

\begin{IEEEbiography}[{\includegraphics[width=1in,height=1.25in,clip,keepaspectratio]{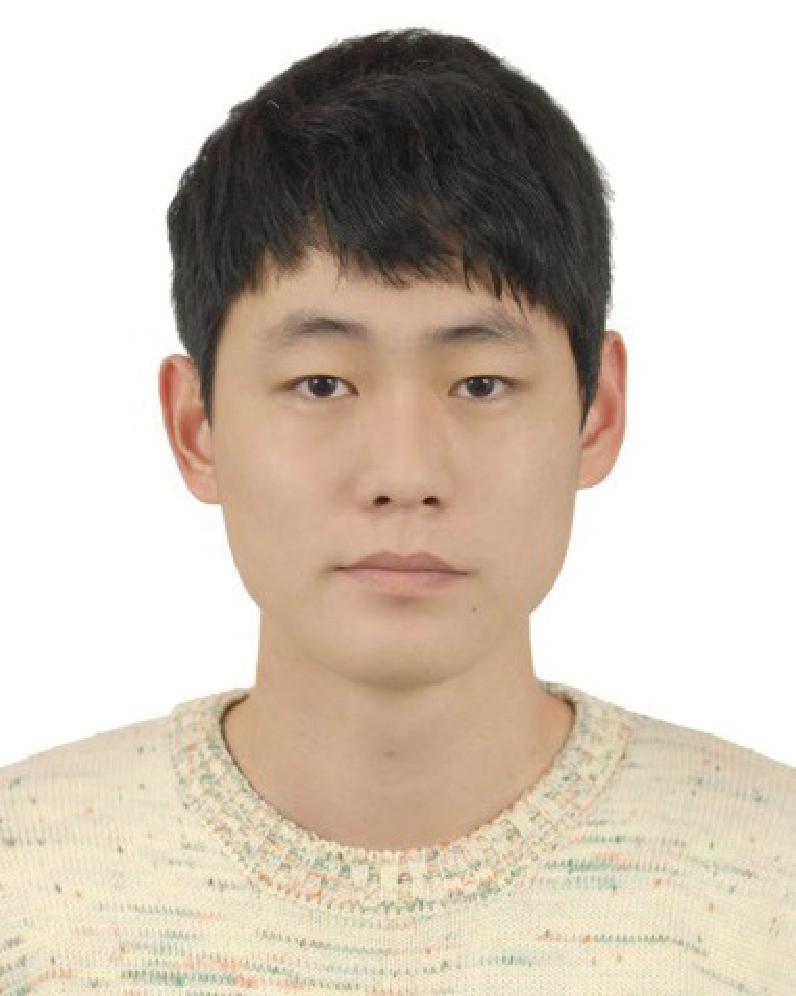}}]{Chaoran Li} received the Bachelor of Information Technology degree from Deakin University Australia in 2018. He is currently working towards the Ph.D. degree at Swinburne University of Technology. His research interests include machine learning, especially in adversarial deep learning.
\end{IEEEbiography}

\vskip 0pt plus -1fil

\begin{IEEEbiography}[{\includegraphics[width=1in,height=1.25in,clip,keepaspectratio]{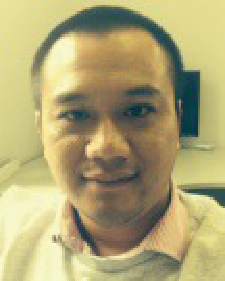}}]{Sheng Wen} received his Ph.D. degree from Deakin University, Australia, in October 2014. Currently he is a senior lecturer in Swinburne University of Technology. He has received over 3 million Australia Dollars funding from both academia and industries since 2014. He is also leading a medium-size research team in cybersecurity area. He has published more than 50 high-quality papers in the last six years in the fields of information security, epidemic modelling and source identification. His representative research outcomes have been mainly published on top journals, such as IEEE Transactions on Computers (TC), IEEE Transactions on Parallel and Distributed Systems (TPDS), IEEE Transactions on Dependable and Secure Computing (TDSC), IEEE Transactions on Information Forensics and Security (TIFS), and IEEE Communication Survey and Tutorials (CST). His research interests include social network analysis and system security.
\end{IEEEbiography}

\vskip 0pt plus -1fil

\begin{IEEEbiography}[{\includegraphics[width=1in,height=1.25in,clip,keepaspectratio]{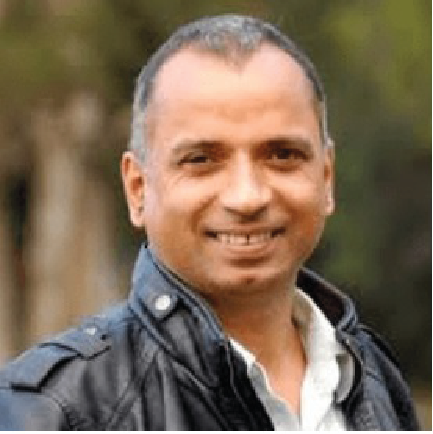}}]{Surya Nepal} Dr Surya Nepal is a Senior Principal Research Scientist at CSIRO Data61. He currently leads the distributed systems security
group. His main research focus is on security, privacy, and trust. He is a member of the editorial boards of IEEE Transactions on Service Computing, ACM Transactions on Internet Technology and Frontiers of Big Data- Security Privacy, and Trust. He is also a theme leader of the Cybersecurity Cooperative Research Centre (CRC), a national initiative in Australia. He holds a conjoint faculty position at UNSW and an honorary professor position at Macquarie University.
\end{IEEEbiography}

\vskip 0pt plus -1fil

\begin{IEEEbiography}[{\includegraphics[width=1in,height=1.25in,clip,keepaspectratio]{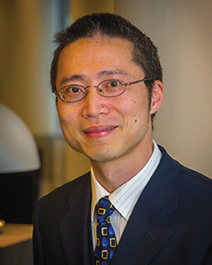}}]{Yang Xiang} received his Ph.D. in Computer Science from Deakin University, Australia. He is currently a full professor and the Dean of Digital Research \& Innovation Capability Platform, Swinburne University of Technology, Australia. His research interests include cyber security, which covers network and system security, data analytics, distributed systems, and networking. In particular, he is currently leading his team developing active defense systems against large-scale distributed network attacks. He is the Chief Investigator of several projects in network and system security, funded by the Australian Research Council (ARC). He has published more than 200 research papers in many international journals and conferences. He served as the Associate Editor of IEEE Transactions on Dependable and Secure Computing, IEEE Internet of Things Journal, IEEE Transactions on Computers, IEEE Transactions on Parallel and Distributed Systems, and the Editor of Journal of Network and Computer Applications. He is the Coordinator, Asia for IEEE Computer Society Technical Committee on Distributed Processing (TCDP). He is currently an IEEE Fellow.
\end{IEEEbiography}
\end{document}